\documentclass[pmlr]{jmlr}

\usepackage{longtable}

\usepackage{booktabs}
\usepackage{siunitx}

\theorembodyfont{\upshape}
\theoremheaderfont{\scshape}
\theorempostheader{:}
\theoremsep{\newline}

\jmlrvolume{1}
\jmlryear{2024}
\jmlrworkshop{AI for Education: Bridging Innovation and Responsibility}

\title{Generating Piano Practice Policy with a Gaussian Process}

\author{\Name{Alexandra Moringen} 
        \Email{alexandra.moringen@uni-greifswald.de}\\
        \addr University of Greifswald
\AND
        \Name{Elad Vromen} 
        \Email{eladvromen@gmail.com}\\
        \addr Tel Aviv University
\AND
        \Name{Helge Ritter}
        \Email{helge@techfak.uni-bielefeld.de}\\
        \addr Bielefeld University
\AND
        \Name{Jason Friedman} 
        \Email{jason@tau.ac.il}\\
        \addr Tel Aviv University}

\begin{document}

\maketitle

\begin{abstract}
A typical process of learning to play a piece on a piano consists of a progression through a series of practice units that focus on individual dimensions of the skill, the so-called practice modes. Practice modes in learning to play music comprise a particularly large set of possibilities, such as hand coordination, posture, articulation, ability to read a music score, correct timing or pitch, etc.  Self-guided practice is known to be suboptimal, and a model that schedules optimal practice to maximize a learner's progress still does not exist.   Because we each learn differently and there are many choices for possible piano practice tasks and methods, the set of practice modes should be dynamically adapted to the human learner, a process typically guided by a teacher. However, having a human teacher guide individual practice is not always feasible since it is time-consuming, expensive, and often unavailable. In this work, we present a modeling framework to guide the human learner through the learning process by choosing the practice modes generated by a policy model. To this end, we present a computational architecture building on a Gaussian process that incorporates  1) the learner state, 2) a policy that selects a suitable practice mode, 3) performance evaluation, and 4) expert knowledge.  The proposed policy model is trained to approximate the expert-learner interaction during a practice session.  In our future work, we will test different Bayesian optimization techniques, e.g., different acquisition functions, and evaluate their effect on the learning progress. 
\end{abstract}

\section{Introduction}
\label{sec:intro}
It can take many years to master complex motor tasks, such as learning to play the piano \citep{furuya_flexibility_2013}. Such learning is typically guided by an expert (a piano teacher). However, access to experts is limited and expensive, and even learners who do learn with an expert are required to practice alone for many hours a week to achieve mastery.

In order to guide or accelerate learning (i.e., maximize performance improvement in the long run), the learner should be presented with practice targets that take into account their skill level, their learning goal, the characteristics of the piece, and the predicted practice utility. The goal of this work is, therefore, to approach the modeling of the structure that underlies this complex scenario in order to approximate the expert-learner interaction. 

One role of a teacher is to assist the learner using scaffolding \citep{moringen_optimizing_2021}. Scaffolding provides temporary guidance for a learner during the learning process \citep{seel_scaffolding_2012}, which can then be gradually removed or faded out. In piano playing, scaffolding can take the form of task simplification, i.e., practicing a particular aspect of performance. For example, this could involve focusing only on pitch (without concern for timing), practicing each hand separately, increasing/decreasing the velocity, or only focusing on timing (by clapping or playing a single note to the rhythm of the piece). The other approach, which we would like to term as \textit{adversarial scaffolding}, is also widely used in training motor tasks and is characterized by artificially increasing the complexity of the task by adding complexity on top \citep{raviv_how_2022}. One extreme technique professional musicians use is to play the piece in reverse order, from the end to the beginning. Teachers use different training approaches for both beginners \citep{thompson_john_2009} and advanced learners (see Cortot's recommended piano exercises for learning to play Chopin \citep{cortot_chopin_1986}). The focus on practicing one or several modalities of the complex skill by increasing or reducing the overall task's complexity is termed practice mode (PM) \citep{moringen_optimizing_2021}.

During practice, we want to optimize the selection of activities the learner performs.  
To do this, we propose a minimalist framework that can be later extended to a more complex scenario.  First, we restrict the actions to two \textbf{practice modes}, timing practice and pitch practice. Second, we capture the \textbf{state of the learner} by recording their performance before and after learning with a given practice mode.  Currently, the state is represented by two error modalities: timing error and pitch error (see a more detailed description in \cite{moringen_optimizing_2021} and Section~\ref{sec:errorCalculations}). We assume utility is a function of the learner state, music piece parameters, and action~\citep{tamir-ostrover_automatic_2022}. 
Based on the ground truth recorded during expert-learner interaction,  we train a Gaussian Process (GP)~\citep{rasmussen_gaussian_2005} to predict the practice utility value. 

A GP has multiple advantages over linear regression, such as the ability to employ acquisition functions for decision-making, the possibility of using different kernel functions, and providing a natural way to quantify uncertainties in prediction. 
Based on the estimated utilities for different practice modes, the long-term goal is to choose the one that corresponds to expert choice.

In this study, we fit a GP into the data acquired from the learner performance and the teacher's decision-making, including the target tempo  and the  practice mode. The goal is to enable a GP model, based on the learner's state, to generate schedules that mimic the guidance of an expert teacher.  By doing that, using the model for guidance should, in the long run, improve the effectiveness of practice, even if the teacher does not guide the learner. We will compare the performance of the GP to a linear regression model to determine whether a GP is necessary for modeling this sort of problem.

\section{Experimental Setting}
Further details about the experimental setting can be found in Appendix~\ref{apd:experimentSetting}.

\paragraph{Participants} The experimental data were collected from 6 novice piano players aged 21 to 28 (who know how to read sheet music, but are at a beginner level). An experienced piano teacher accompanied them. The same teacher was present for all sessions. The study received ethical approval from the Tel Aviv University Institutional Review Board (IRB) and participants provided written informed consent before the experiment. The participants received payment for their time. Each session was approximately 1 hour. 

\begin{figure}[!htb]
\begin{center}
\includegraphics[width=0.7\textwidth]{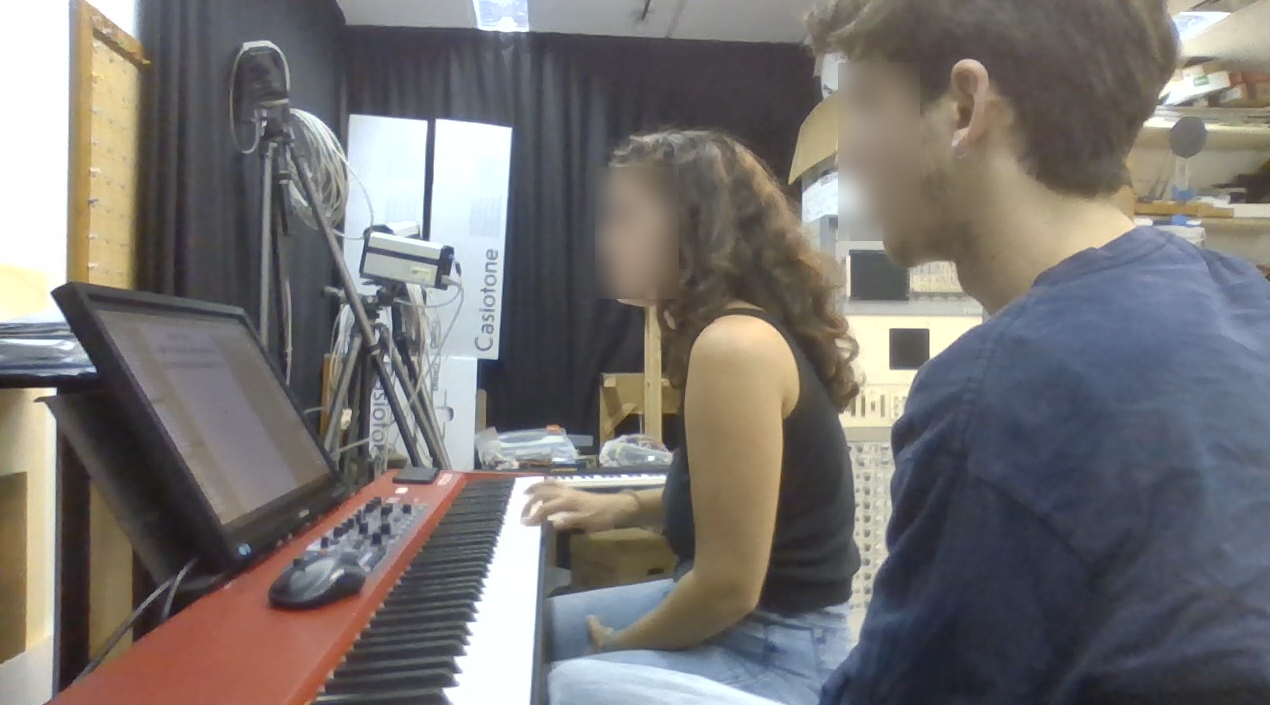}
\end{center}
\caption{\label{figure:blurred_faces} The experimental setup is shown in this photo. The learner sits at the piano (Nord piano 4) with a computer monitor showing the notes to play. Using the Python interface, the teacher selects which practice mode to play and when to change the piece.}
\end{figure}

\paragraph{Equipment} The learner sat at an electric piano (Nord Piano 4), and the practice tasks that the teacher selected were displayed on a computer monitor placed on the piano notes stand. The MIDI data produced by the learners was transferred to the computer for later analysis. Figure~\ref{figure:blurred_faces} shows a photo of the setup. 

\paragraph{Selection of piece to play} The piece to play was selected by the teacher, who chose a piece that he thought would be suitable and preferably a bit challenging for the student. Sometimes, when the pieces were too difficult or easy, the piece was changed to a different one after two or three practices. All pieces were for the right hand only. 

\begin{figure}[!htb]
\begin{center}
\includegraphics[width=0.9\textwidth]{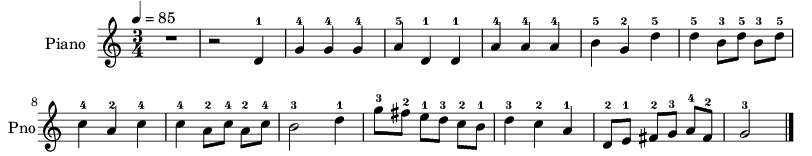}\\
\vspace{1.2cm}
\includegraphics[width=0.9\textwidth]{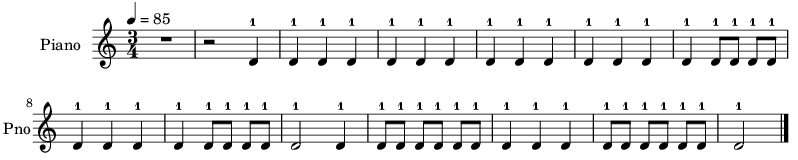}
\end{center}
\caption{\label{fig:practicemodes}The upper picture shows an example of one of the pieces to be played. The lower picture shows the selection of a timing practice mode (PM) - the pitch of all notes is uniform, allowing the learner to focus on the timing.}
\end{figure}

\paragraph{Practice Modes} In the experiment, only two practice modes were used: timing and pitch. An example of a practice mode can be found in Figure~\ref{fig:practicemodes}. 

The \textbf{timing practice mode} reduces the pitch of all notes to a single pitch and, by doing so, enables the learner to focus on rhythm and timing only, getting rid of the information about the pitch. The \textbf{pitch practice mode} allows the learner to practice only the pitch of the music piece and not pay attention to the timing of the notes. 
To this end, the participants were instructed to focus on the pitch and not the rhythm. A metronome was used in all trials (with the tempo set by the teacher) except in the pitch practice mode.

\paragraph{Teacher role} The teacher defined each practice unit, defined by the music piece, the practice mode, and the tempo (in beats per minute - BPM). The original piece was played by the learner before and after each practice mode to allow calculation of the utility of practice, defined as the difference in errors before and after practice. After the full piece was played, the teacher determined again which practice mode would then be likely to be most beneficial.  The teacher also decided when it was time to move to a different piece.  From all the experimental sessions, we acquired 121 tuples of performances, consisting of performance before the teacher-selected practice and performance after the practice mode. 

\section{Methods: Models and Features}
\subsection{Calculation of features from user data}
\label{sec:errorCalculations}
This work is inspired by a previous study~\citep{moringen_optimizing_2021}. We provide a brief overview below of the calculated features that we used for modeling based on the recorded performance of the learners: pitch error, timing error, and utility of a practice mode. The raw data of learners playing the piano is recorded in MIDI, a protocol that sends out a message every time a note is pressed or released. To simplify the modeling, in this work we do not use MIDI directly, but extract learner performance features based on the ground truth, i.e., correct performance of the piece. \\ 
\textbf{Pitch error} is calculated as follows for a given piece: $error_{pitch}(n): = \frac {\sum_{i=1}^T d(n_i) f(n_i)}{\sum_{i=1}^T d(n_i)},$ 
where $n:= (n_1, \ldots, n_T)$ is the sequence of notes in the piece; 
$f$ is a function that returns whether the note pitch was played correctly (0 if correct; 1 if incorrect);  $d$ is a function that returns the duration of the given note (e.g., 1.0 for a quarter note and 2.0 for a half note).\\
\textbf{Timing error} is calculated as: $error_{timing}(n): = \frac{1}{m} \sum_{i=1}^m \min\{1, t(n_i)\},$ 
where $t$ is a function that returns the timing offset for the given note as the offset between the time the note was played and the correct time. $n:=(n_1, \ldots, n_m)$ is the sequence of notes played (excluding missed notes). All error measures were calculated based on performance of the original pieces. This work did not analyze errors during practice with the practice modes. 

A central feature of our method is the utility $u$ of practice, which measures how effective a practice mode is.  
To quantify the \textbf{utility} $u$  of a practice unit, we measure
the error of the learner’s performance, playing the whole music piece before practicing and after practicing in a certain practice mode (${pre}$ and ${post}$ denote before and after the practice, respectively):   
\begin{equation}
\label{eq:utility}
    u (a, u_\mu) = (1-a) (p_{pre} - p_{post}) + a(t_{pre} - t_{post}) - u_{\mu},
\end{equation}
where $p$ and $t$ are the pitch and timing errors, respectively, the weight $a$ 
represents the contribution of each modality’s improvement to the overall utility value, and $u_{\mu}$ is the mean utility.
Note that both $u_{\mu}$ and $a$ are unknown.  In the next section, we describe the Bayesian optimization that we use to find the best values for these parameters.  The thinking behind optimizing these parameters is to understand how the teacher combines both of these measures to estimate the utility of a given practice mode. 

\section{Gaussian process and the scaffolding method}
 The GP was constructed to align the model's prediction to the teacher's selection.  It is formalized as a utility predictor, given the state of the learner before practice, practice mode, and speed of practice: 
 \begin{equation}  
 g: (p_{pre}, t_{pre}, PM, BPM) \rightarrow u(a, u_\mu),  
 \end{equation}  
 where $u$ is calculated as in Eq.~\ref{eq:utility}.  
  The left side of the above equation corresponds to the user state and the method of practice, while the right side is trained to reflect the usefulness estimation of the corresponding practice by the teacher. 
  \textbf{We optimize the hyperparameters $a$, $u_\mu$, and the GP kernel so that by comparing the utility estimators for different PMs, the highest estimated utility corresponds to the practice chosen by the teacher.} The objective function we employ in the optimization process uses the utility estimates but only to calculate whether the maximum of the predicted utility measures corresponds to the practice mode selected by the teacher: 
  $PM_{teacher} == arg\,max \{ g(PM_{timing}), g(PM_{pitch})\}$. This approach does not consider whether the GP model makes correct predictions of the utility value. 
  Note that the current implementation  of the above-mentioned scaffolding algorithm makes a strong simplification by assuming that the teacher maximizes the utility of a practice mode based on both values, without any consideration of the practice history. 
  In future work, we plan to additionally employ the uncertainty of the utility estimator that a GP also provides. It is an open research question, how to integrate the practice context into a computational practice policy generator. 
At the moment we can answer the following questions based on the current learner state only: ``For my current level of errors, what tempo should I select for timing practice of the piece?'' or ``For my current level of errors, what is the best PM to choose, and at what speed to practice?''
 
The kernels examined here were the rational quadratic kernel (RatQuad), radial basis function (RBF), and Mat\'ern with $\nu =5/2$~
 \citep{rasmussen_gaussian_2005}. The hyperparameters we optimized explicitly with Bayesian optimization were  $a$ and $u_\mu$. The other parameters, multidimensional lengthscale, variance and $\alpha$ (in the case of RatQuad) were optimized with the automatic relevance determination (ARD) using the GPyOpt libary\footnote{\url{https://sheffieldml.github.io/GPyOpt/}}  

\section{Methods - Linear regression}
We compared the GP with linear regression to test whether linear regression is sufficient for predicting which PM to use. We fit a linear regression model to the training data to predict the utility $u$ (Equation~\ref{eq:utility}) as a function of the pitch and timing error before the practice, the practice mode used, and their interactions:
\begin{equation}
u = b_1 + b_2 \cdot PM + b_3 \cdot t_{pre} + b_4 \cdot p_{pre} + b_5 \cdot PM \cdot t_{pre} + b_6 \cdot PM \cdot p_{pre}
\end{equation}
where $b = \begin{bmatrix} b_1 & b_2 & b_3 & b_4 & b_5 & b_6\\ \end{bmatrix}$ are the regression parameters fit to the data.

The linear regression was trained to predict the utility on data for all subjects combined. As the utility function has one free parameter ($a$ in Equation~\ref{eq:utility}), we ran multiple linear regressions (using a grid search to test exhaustively values from 0 to 1) to find the one that most accurately predicts the teacher-selected activity assuming that the teacher selects the activity with the greatest utility. We also tested whether including tempo (BPM) improved the prediction accuracy. In addition, we tested whether directly predicting the teacher's selection (i.e., which practice mode - $PM$) would be more successful using logistic regression:
\begin{equation}
PM = b_1 + b_2 \cdot t_{pre} + b_3 \cdot p_{pre}
\end{equation}
\section{Results}
\subsection{Gaussian process}
\begin{figure}[htb]
\begin{center}
\includegraphics[width=0.4\linewidth]{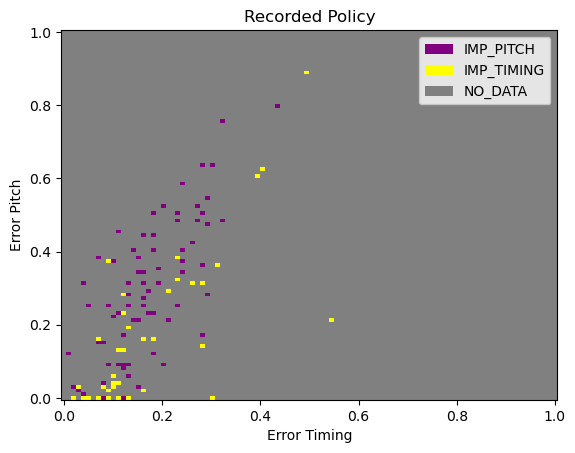}
\includegraphics[width=0.4\linewidth]{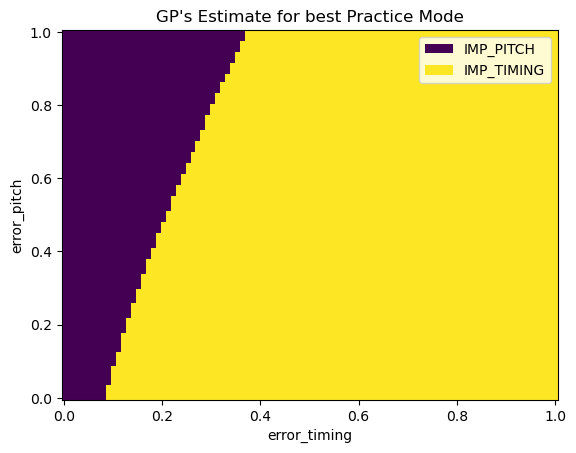}\\
\includegraphics[width=0.4\linewidth]{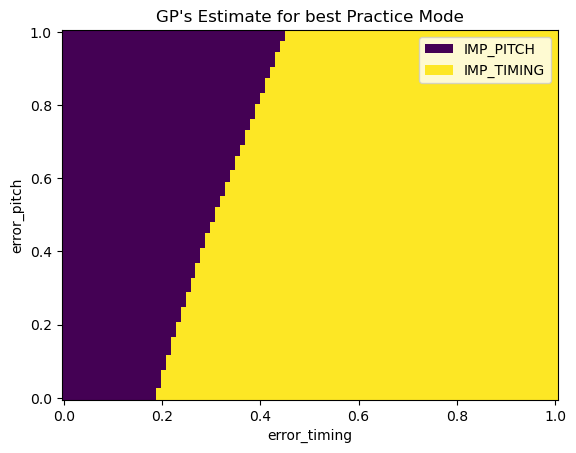}
\includegraphics[width=0.4\linewidth]{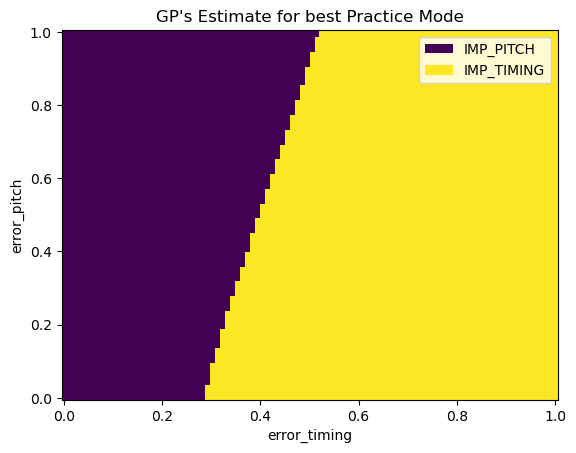}
\end{center}
\caption{\label{fig:recordedpolicy} (Upper left) An illustration of data points recorded during expert-guided piano practice session for all learners and pieces. $x$-axis denotes timing error; $y$-axis denotes pitch error before a PM (pitch practice in violet, timing practice in yellow). (Upper right, lower panels) Policies generated by a GP RatQuad kernel, BMP=50, 80 and 100. For slower tempos, the model's policy predicts a higher utility of timing practice; for higher tempos - a higher utility of pitch practice.}
\end{figure}

Figure~\ref{fig:recordedpolicy} shows an example of the practice policy provided by the teacher and policies generated by maximizing the estimated practice utilities predicted by a trained GP for BPM=50, 80, and 100, using a RatQuad kernel. After 50 iterations of Bayesian optimization of $a$ and $u_\mu$, the scaffolding algorithm on average correctly predicts 70\% of the practice modes on test data. This computation takes about a minute on a regular PC.  Further Bayesian optimization iterations result in a slight increase of accuracy, 
yielding 75\% accuracy. 
  
\subsection{Linear Regression}
The linear regression was able to predict reasonably well the utility based on the error measures before the practice ($R^2=0.357, p<0.001$).
We found that selecting the activity based on the maximum utility was closest to the ground truth (from the piano teacher) when $a=[0,0.13]$ (i.e., any $a$ value chosen in this range was equivalent). We note that this includes the value of $a=0$, i.e., only using the pitch data.
However, the accuracy of the linear regression was poor (only $38.0\%$), i.e., less than chance.

When using logistic regression to predict the teacher's selection directly, the accuracy was somewhat higher ($69.4\%$); this increases to $70.2\%$ if BPM is included. In this case, the equation was:
\begin{equation}
PM = -0.283 + 8.124 \cdot t_{pre} - 6.734 \cdot p_{pre}
\end{equation}
where the timing practice model was selected if $PM > 0.5$.

\section{Discussion}
The GP produced a mapping between errors and desired practice modes that makes sense (see Figure~\ref{fig:recordedpolicy}), while the accuracy of both techniques was similar (approximately 70\% for the baseline model and 75\% as the best result achieved by the GP model). Using more error measures may show a larger difference between the techniques because it is likely that the relationship between predictor variables and the utility is non-linear \citep{schulz_tutorial_2018}. 

The model has a greater tolerance for small timing errors than pitch errors - when the timing error is low, the pitch PM is always selected. This may be because variability in timing is a typical feature of playing \citep{vugt_fingers_2012}, whereas any variability in pitch (which key is pressed) is considered an error.

Future studies will compare using the GP to provide learners with the next PM to a control group who practice with randomly scheduled PMs. The goal is to test whether this use of GP  accelerates motor learning of the piano.  Another research direction is to use different types of acquisition functions and investigate how similar the schedule generated by the teacher and by the acquisition functions are, and whether it makes sense for a learner to practice within a Bayesian optimization loop. A further challenge is to find an approach to modeling of the practice context within one session.   

\section{Data availability}
The code for data collection and analysis, and the collected data are \href{https://github.com/JasonFriedman/piano_GP/}{available}.

\acks{This study was funded by the German-Israeli Foundation for Scientific Research \& Development.}
\newpage
\bibliography{refs,zotero}
\newpage
\appendix
\section{Experimental setting}\label{apd:experimentSetting}
We will conduct an experiment to develop an artificial intelligence algorithm that will assist in learning piano playing. The algorithm will provide accurate recommendations about what the ideal exercise is for the learner based on their mistakes while playing at the current moment. The current experiment is intended to provide a basic ground truth for training the model. In this experiment, the expert will perform the algorithm's decision-making process in its place and will guide the choice of exercises for the learner based on her playing at the current moment. As usual in preliminary computational models, it's important to clarify that this is currently a limited model. The model measures only two errors out of the learning process - pitch and timing errors. We aim to create a successful model based on these two, and later, this can be expanded to include other types of errors. 

The current experiment is performed iteratively. First, the expert must choose a piece that suits the learner and is at an appropriate pace. Then, the learner will play the piece in its entirety. The expert will assess her playing and recommend one of two training exercises, according to his professional discretion: Pitch: playing only the correct notes, without regard to rhythm. Timing: playing according to rhythm only, without regard to which note (key) to press. According to the choice, input will be entered into the system, and the software will show the learner the chosen exercise. In this experiment, we will perform each practice once. Afterward, the learner plays the full piece again to assess the improvement in her error. The process will be performed iteratively until the errors in the piece are sufficiently small, in the expert's opinion. Then, we will move on to the next piece.
\end{document}